\documentclass{article}

\usepackage{times}
\usepackage{graphicx} 

\usepackage{natbib}

\usepackage{algorithm}
\usepackage{algorithmic}

\usepackage{amsmath}
\usepackage{amsfonts}
\usepackage{caption, subcaption}
\usepackage{hyperref}

\usepackage[accepted]{whi2016}

\icmltitlerunning{ACDC: $\alpha$-Carving Decision Chain for Risk Stratification}

\begin{document} 

\setlength{\abovedisplayskip}{5pt}
\setlength{\belowdisplayskip}{5pt}

\twocolumn[
\icmltitle{ACDC: $\alpha$-Carving Decision Chain for Risk Stratification}

\icmlauthor{Yubin Park}{yubin@accordionhealth.com}
\icmladdress{Accordion Health, Inc.,
            4200 N. Lamar Blvd., Austin, TX 78756 USA}
\icmlauthor{Joyce Ho}{joyce.c.ho@emory.edu}
\icmladdress{Emory University,
            400 Dowman Dr, Atlanta, GA 30322 USA}
\icmlauthor{Joydeep Ghosh}{jghosh@utexas.edu}
\icmladdress{The University of Texas at Austin,
            Austin, TX 78712 USA}            

\icmlkeywords{Decision tree, $\alpha$-divergence}

\vskip 0.3in
]

\begin{abstract} 
In many healthcare settings, intuitive decision rules for risk stratification can help effective hospital resource allocation.
This paper introduces a novel variant of decision tree algorithms that produces a chain of decisions, not a general tree. 
Our algorithm, $\alpha$-Carving Decision Chain (ACDC), sequentially carves out ``pure'' subsets of the majority class examples.
The resulting chain of decision rules yields a pure subset of the minority class examples.
Our approach is particularly effective in exploring large and class-imbalanced health datasets.
Moreover, ACDC provides an interactive interpretation in conjunction with visual performance metrics such as Receiver Operating Characteristics curve and Lift chart.
\end{abstract}

\section{Introduction}

Data analytics has emerged as a vehicle for improving healthcare \cite{Meier2013} due to the rapidly increasing prevalence of electronic health records (EHRs) and federal incentives for meaningful use of EHRs.
Data-driven approaches have provided insights into diagnoses and prognoses, as well as assisting the development of cost-effective treatment and management programs.
However, there are two key challenges to the development of health data analytic algorithms: 1) noisy and multiple data sources issues, and 2) interpretability issues.

A decision tree is a popular data analytics and exploratory tool in medicine \cite{Podgorelec2002,Lucas1998,Yoo2012} because it is readily interpretable.
Decision tree algorithms generate tree-structured classification rules, which is written as a series of conjunctions and disjunctions of the features.
Decision trees can produce either output scores (a positive class ratio from a tree node) or binary classes (0/1).
Not only are the classification rules readily interpretable by humans, but also the algorithms naturally handle categorical and missing data.
Therefore, various decision trees have been applied to build effective risk stratification strategies \cite{Fonarow2005,Chang2005,Goto2009}.

We believe that decision trees for risk stratification can be improved from two aspects.
First, many existing approaches to class-imbalance problems typically rely on either heuristics or domain knowledge \cite{Chawla2002,Japkowicz2000,Domingos1999}.
Although such treatments may be effective in some applications, many of them are \textit{post-hoc}; the splitting mechanism of a decision tree usually remains invariant.
Second, even the logical rules from a decision tree can be overly complex, especially with class-imbalanced data.
Furthermore, the conceptual gap between decision thresholds and decision rules complicates interpretation on visual performance metrics such as the receiving operating character (ROC) curve and the lift chart.

\begin{figure}
\center
\begin{subfigure}[b]{0.3\linewidth}
\centering
\includegraphics[width=1\linewidth]{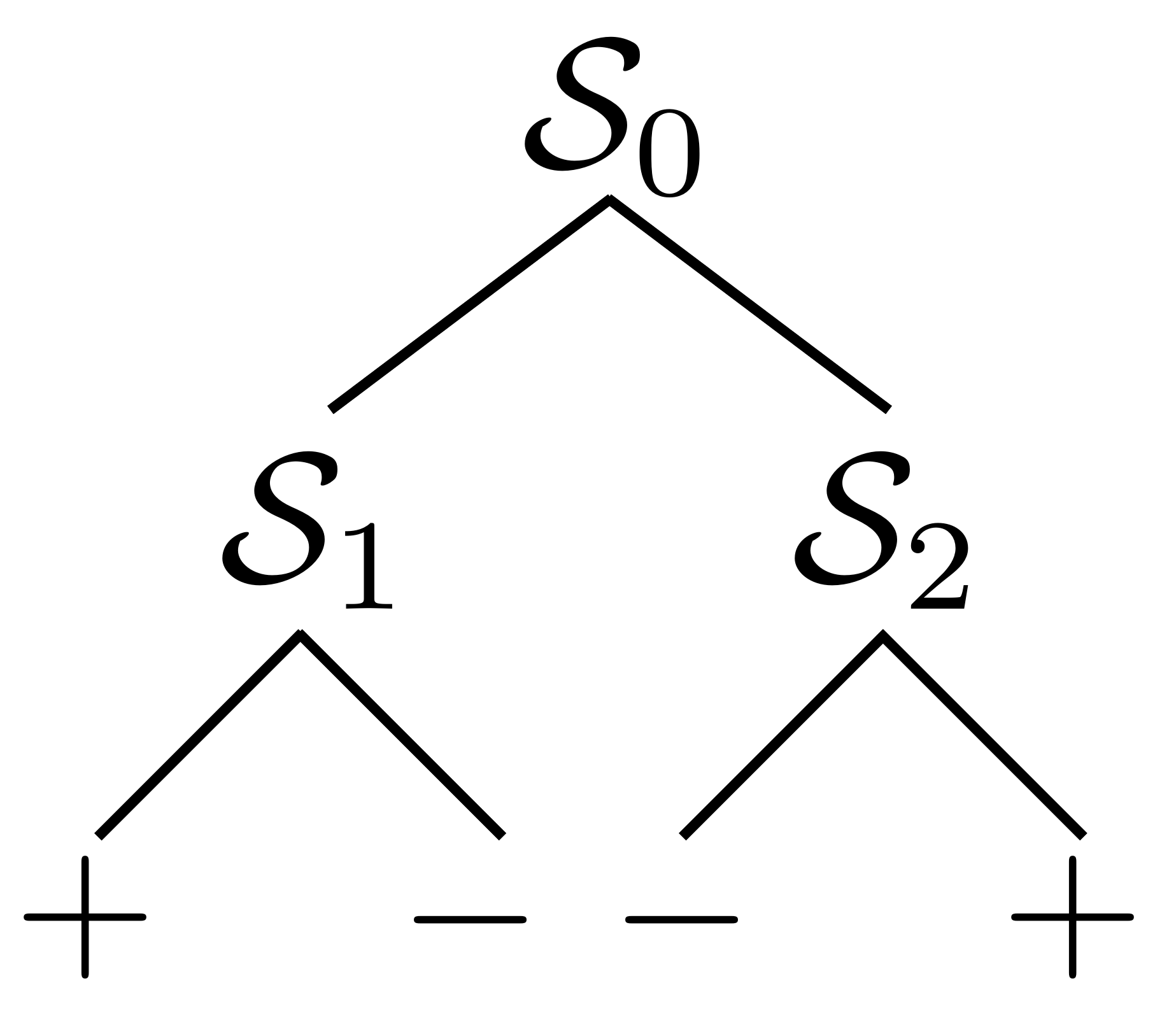}
\caption{A decision tree}
\label{fig:dt}
\end{subfigure}
\quad\quad\quad
\begin{subfigure}[b]{0.32\linewidth}
\centering
\includegraphics[width=1\linewidth]{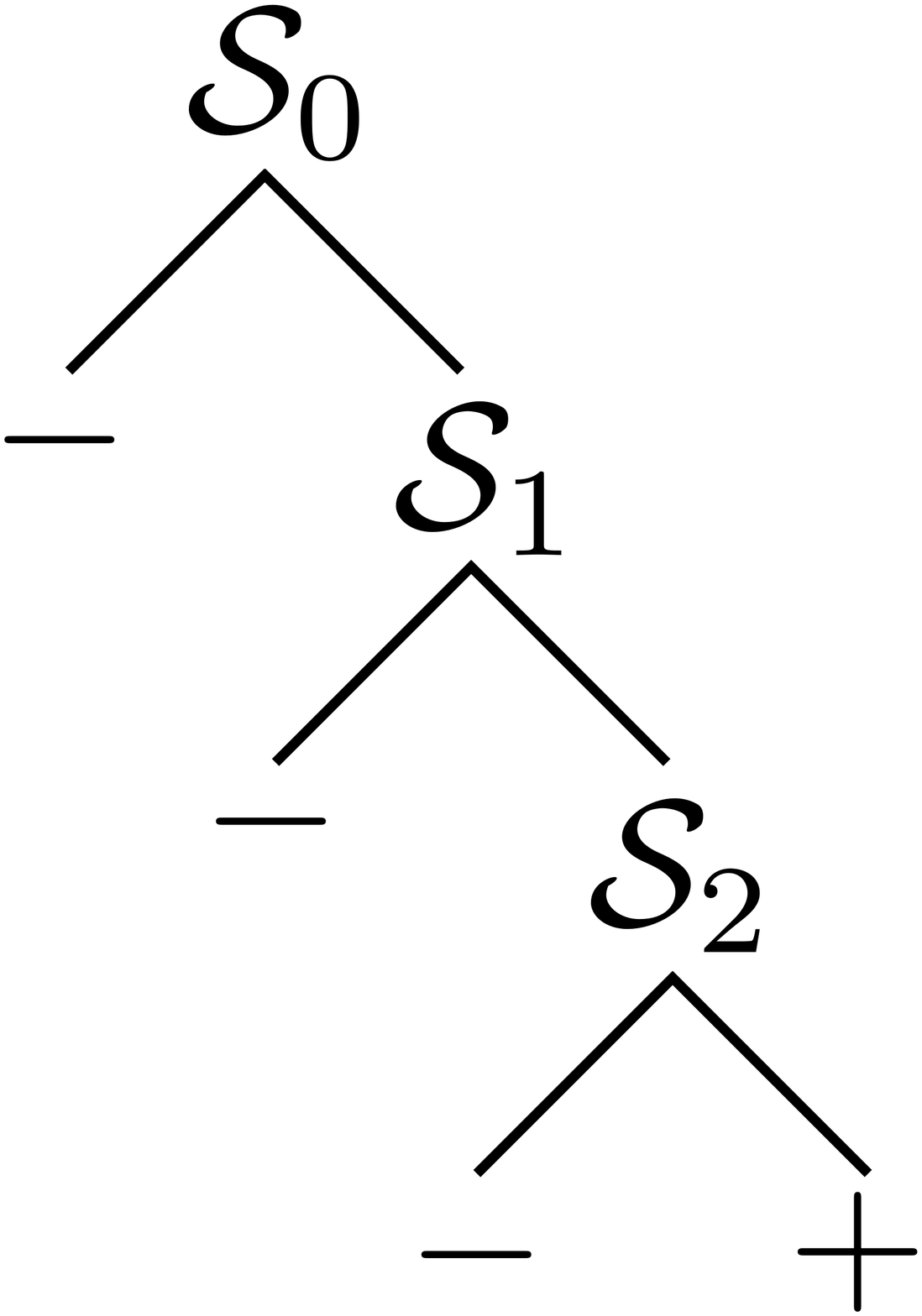}
\caption{A decision chain}
\label{fig:dc}
\end{subfigure}
\vspace{-10pt}
\caption{The difference between a decision tree and a decision chain. 
The output partitions of a decision tree are not ordered.
On the other hand, the outputs of a decision chain satisfy the monotonic risk condition.}\label{fig:diff}
\end{figure}

We propose {$\boldsymbol{\alpha}$-Carving Decision Chain (ACDC)}, a novel variant of decision tree algorithms, that produces \textit{a chain of decisions} rather than a tree.
Conceptually, ACDC is a sequential series of rules, applied one after another, where the ratio of positive class increases over the sequence of decision rules (i.e. \textit{monotonic risk condition}). 
Figure \ref{fig:diff} presents a comparison between a decision tree and a decision chain.
Thus, the decision order creates a noticeable difference in the number of distinct rules.
The idea of constructing a decision chain has been recently explored using Bayesian models \cite{WangRu15,HongyuYang16}.
These models have shown promising results in terms of interpretability and predictive performance. 
ACDC can be viewed as an alternative to such models.
Our greedy chain growing strategy is particularly well-suited when exploring large and class-imbalanced datasets. 

ACDC is based on the $\alpha$-Tree framework \citep{Park2012:tkde} developed for imbalanced classification problems.
The key idea of ACDC is to sequentially carving out ``pure'' subsets of majority examples (the definition of purity is given in Section~\ref{sec:afactor}).
Each subsequent step in ACDC yields a higher minority class ratio than the previous steps.
The step-wise approach allows our algorithm to scale readily to large data and handle class-imbalance problems.
We demonstrate ACDC on two real health datasets and show that our algorithm produces outputs that are concise and interpretable with respect to visual performance metrics, and achieves predictive performance comparable to traditional decision trees.
ACDC can be used for various healthcare applications, including (i) symptom development mining, (ii) step-wise risk-level identification, (iii) early characterization of easily identifiable pure subsets, and (iv) decision thresholds determination with decision rules.

\section{ACDC: $\alpha$-Carving Decision Chain}\label{sec:acdc}

ACDC is motivated by the need to interpret large and class-imbalanced healthcare datasets. 
While exploring several such datasets, we have frequently observed that negative class examples can be easily carved out with simple rules. 
We initially attempted to apply various heuristics, such as cost-weighted scoring functions \cite{Buja2001,Buja2005}, to construct such rules, but realized that this approach does not scale with different types of datasets. 
Every time we encountered a new dataset, we needed new domain knowledge to filter out negative class examples.
Thus, we developed ACDC,  a novel variant of decision tree algorithms.
ACDC produces a chain of decisions by sequentially carving out pure subsets of the majority class ($Y = 0$) examples, and provides a systematic approach to construct such filtering rules.

To achieve this goal, we introduce (i) a new criterion for selecting a splitting feature, (ii) its implication, and then (iii) using the criteria, a simple dynamic strategy to grow a one-sided tree.

\subsection{Selecting a Splitting Feature}
The splitting criterion for the $\alpha$-Tree \cite{Park2012:tkde} selects the feature with the maximum $\alpha$-divergence \cite{Amari2007,Cichocki2010} between the following two distributions:
\begin{align*}
\underbrace{\text{P}(X_i,Y)}_{\text{Actual distribution}} \longleftrightarrow \underbrace{\text{P}(X_i)\text{P}(Y)}_{\text{Reference distribution}}
\end{align*}
The reference distribution is set to the product of the two marginals; if both are independent then the reference distribution is equivalent to the joint distribution.
In other words, $\alpha$-Tree selects a splitting feature that maximizes the dissimilarity between the joint and marginal distributions.

Although the $\alpha$-Tree criterion is conceptually simple, it is difficult to control and analyze.
Instead, we simplify the reference distribution as follows:
\begin{align*}
&\text{U}(X_i,Y) = \text{U}(Y \mid X_i) \text{U}(X_i)\\
&\text{s.t}\quad \text{U}(Y \mid X_i) = \frac{1}{2} \quad\text{and}\quad \text{U}(X_i) = \text{P}(X_i)
\end{align*}
The reference distribution $\text{U}(X_i,Y)$ changes with respect to a feature $X_i$.
Integrating $\text{U}(X_i,Y)$ over $Y$ yields a distribution that is the same as the marginal of $X_i$.
Furthermore, given $X_i$, the reference distribution becomes the uniform distribution as it has no information on $Y$.

We modify the $\alpha$-divergence criterion to select the feature that provides the maximum distance between $\text{P}(X_i,Y)$ and $\text{U}(X_i,Y)$.
Therefore, the ACDC-criterion is the following:
\begin{align*}
\arg\max_{i}  \frac{1}{\alpha(1-\alpha)} (1 - \frac{1}{2}\sum_{x_i, y}\text{P}(x_i)(2\text{P}( y \mid x_i))^{\alpha})
\end{align*}

This particular choice of the reference distribution may appear somewhat contrived.
However, the reference distribution automatically captures the splitting criteria of both C4.5 and CART as special cases.
From the ACDC-criterion, we can obtain the information gain criterion (C4.5) by setting $\alpha=1$ and the Gini criterion (CART) by setting $\alpha=2$.
For example, using $\alpha \rightarrow 1$ and L'H\^{o}pital's rule, we can obtain the information gain criterion.

\subsection{Meaning of ACDC-criterion}\label{sec:afactor}
We define a new quantity $\text{A}(p, \alpha)$, the $\alpha$-zooming factor (az.factor), as follows: 

\begin{align*}
\text{A}(p, \alpha) = p^\alpha + (1-p)^\alpha
\end{align*}

The az.factor can have different interpretations:
\begin{itemize}
\setlength\itemsep{0.05em}
\item $\|  (p, 1-p) \|_\alpha^\alpha$, where $\| \cdot \|_\alpha$ is $L^\alpha$ norm
\item a generalized entropy index of $\text{P}(Y)=(p, 1-p)$ in economics literature \cite{Aman1998}
\item a generalized diversity index of $\text{P}(Y)=(p, 1-p)$ in ecological literature  \cite{Simpson1949,Moreno2010,Lou2006}
\end{itemize}
In this paper, we simply use az.factor as a parametrized purity measure and are more interested in its functional role in the ACDC-splitting criterion.

Under the condition $\alpha > 1$, we can rearrange the terms to obtain:
{\small
\begin{align*}
&\max_{X_i} D_{\alpha} ( \text{P}(X_i,Y) \| \text{U}(X_i,Y) ) \\
&= \max_{X_i} \text{A}(\text{PPV}_i,\alpha) \underbrace{\text{P}(X_i=1)}_{\text{Balance term 1}}+\text{A}(\text{NPV}_i,\alpha) \underbrace{\text{P}(X_i=0)}_{\text{Balance term 2}}
\end{align*}
}
where PPV and NPV represent positive and negative predictive values, respectively.
Notice that $\alpha$ emphasizes or \emph{zooms} on these values: PPV and NPV.
As $\alpha$ increases, the splitting criterion prefers higher $\text{P}(Y \mid X)$:
\begin{itemize}
\setlength\itemsep{0.05em}
\item $\alpha \uparrow$: more focus on the PPV and NPV terms
\item $\alpha \downarrow$: more focus on the balance terms
\end{itemize}
Therefore, lower values of $\alpha$ result in more balanced splits, and higher values of $\alpha$ provide very sharp PPV and NPV values (i.e. a pure subset of either the majority or the minority classes).

\subsection{Growing a Decision Chain}\label{sec:algorithm}

Our strategy to build a monotonic decision chain is to gradually decrease the value of $\alpha$.
This is motivated by the following two observations.
First, at each subsequent stage, we have a smaller number of samples.
To prevent biased splits, $\alpha$ should be appropriately adjusted to the current sample size.
Second, as a chain grows, we have a more balanced class ratio at each stage.
If $\alpha$ remains too high, then both PPV and NPV terms numerically have little effect.

We introduce an $\alpha$-carving strategy to adjust $\alpha$ accordingly. 
At each stage of a decision chain, we set $\alpha$ as follows:
\begin{align}
\text{Find}\quad \alpha\quad \text{s.t.}\quad \nu =\left. \frac{\partial \text{A}(\omega, \alpha) }{\partial \omega} \right\vert_{\omega = \omega_y} \notag
\end{align}
where $\nu$ is a predefined velocity parameter, and $\omega_y$ is defined as $\max (\text{P}(Y), 1-\text{P}(Y)) \label{eq:omega-y}$.
As the decision chain builds up, the value of $\alpha$ decreases.

The overall steps of the ACDC algorithm is as follows:
\begin{enumerate}
\setlength\itemsep{0.05em}
\item Set the value of $\nu$
\item Find an appropriate $\alpha$
\item Find a feasible set of splitting variables that satisfy the monotonic risk condition
\item Find a splitting variable from the feasible splitting variable set
\item Discard the majority class examples
\item Repeat from Step 2
\end{enumerate}
Note that ACDC grows only one branch unlike decision trees.
The parameter $\nu$ controls the size of the chain.
A low $\nu$ typically results in a large $\alpha$ and obtains chains that tend to be longer with small-sized partitions.
On the other hand, a large $\nu$ produces a shorter chain with big partitions.

\section{Experimental Results}\label{sec:experiments}

We provide the experimental results of ACDC on MIMIC-II database \cite{Saeed2011} focusing on two different conditions (septic shock and asystole).
For each condition, we will compare the performance with C4.5 and CART and other kinds of alpha trees, show how the cutting plane changes with different values of $\alpha$, and display rule-annotated ROC and Lift charts resulting from ACDC.

The MIMIC-II database is one of the largest publicly available clinical databases.
The database contains more than 30K patients and 40K ICU admission records.
For this paper, we concentrate on two subsets of the database, specifically 1) patients with systemic inflammatory response syndrome (SIRS) for septic shock prediction, and 2) patients with or without cardiac arrests for asystole prediction.
The features are derived primarily from non-invasive clinical measurements and include blood pressure (systolic and diastolic measurements), body temperature, heart rate, respiratory rate, and pulse oximetry. 
For each measurement, we use the last observed measurement and three additional sets of derived features: max, min, average values within the last 12 hours.

\textbf{Septic Shock.}
We first illustrate the results from the septic shock dataset.
Septic shock is defined as ``sepsis-induced hypotension, persisting despite adequate fluid resuscitation, along with the presence of hypo perfusion abnormalities or organ dysfunction'' \cite{Bone1992}.
The time of septic shock onset was defined using the criteria outlined in a recent work on septic shock prediction \cite{Ho:2012tu}.
For this subset, there is a total of 1359 patients with 213 transitioning to septic shock.
We use ACDC and decision trees to predict if a patient will enter septic shock 1 hour prior to shock onset.

\begin{figure}
\centering
\begin{subfigure}[b]{0.45\linewidth}
\includegraphics[width=1\linewidth]{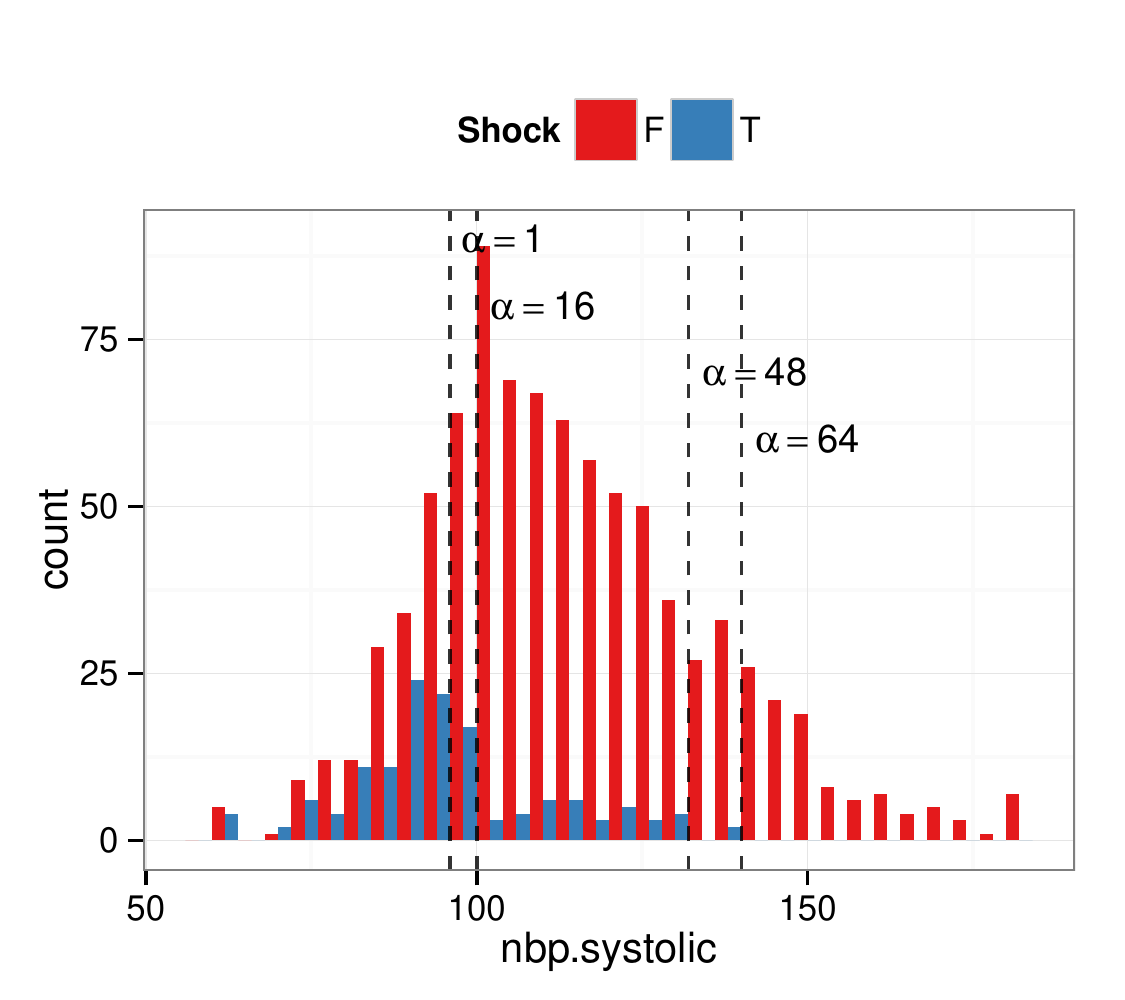}
\caption{Cutting planes w.r.t. $\alpha$}
\end{subfigure}
\begin{subfigure}[b]{0.45\linewidth}
\includegraphics[width=1\linewidth]{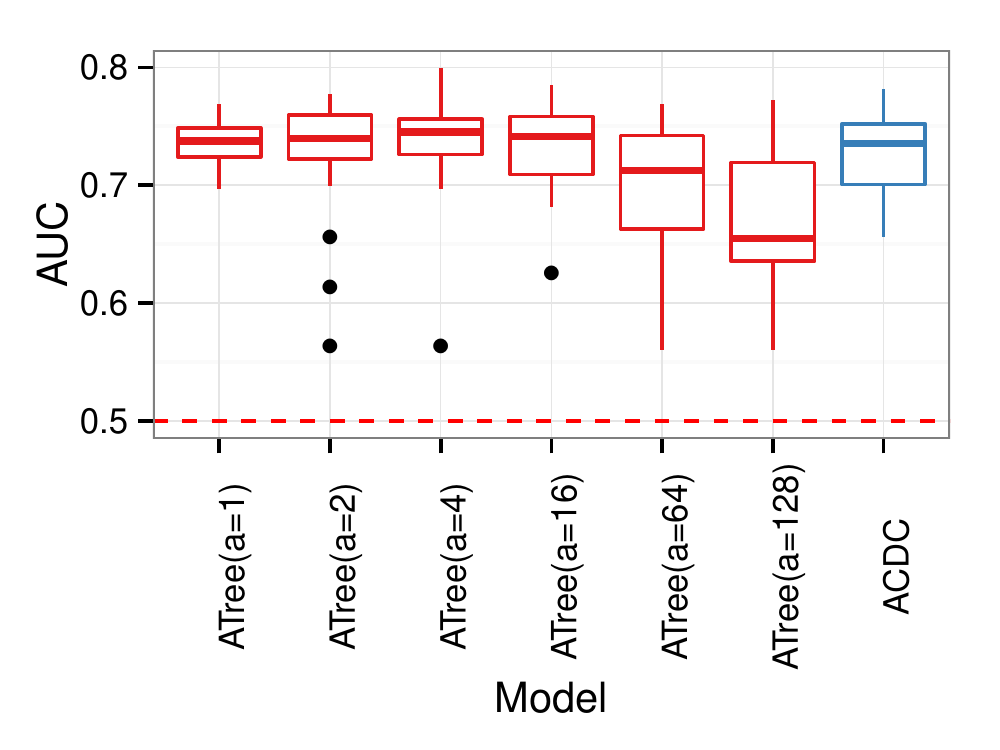}
\caption{AUROCs on hold-out sets}
\end{subfigure}
\vspace{-10pt}
\caption{(a) Different decision cuts with different $\alpha$ values. 
The cut with $\alpha=64$ provides an extremely pure subset of non-septic shock patients. 
(b) The performance of ACDC is comparable to that of $\alpha$-Trees with $\alpha=1$ (C4.5) and $\alpha=2$ (CART). }\label{fig:septic_auc}
\end{figure}

Figure~\ref{fig:septic_auc} (a) shows the first cuts of decision trees with the data collected 1hour before septic shock.
The information gain criterion ($\alpha=1$) selects the cut with systolic=96 mmHg, which almost coincides with the definition of the septic shock.
However, this cut results in a large portion of false negatives.
On the other hand, high values of $\alpha$ reduce such false negatives.
The resultant cuts are rather conservative (smaller number of patients are classified as negatives) but produce a pure subset of non-septic shock patients.

\begin{figure}
\centering
\begin{subfigure}[b]{0.45\linewidth}
\includegraphics[width=\linewidth]{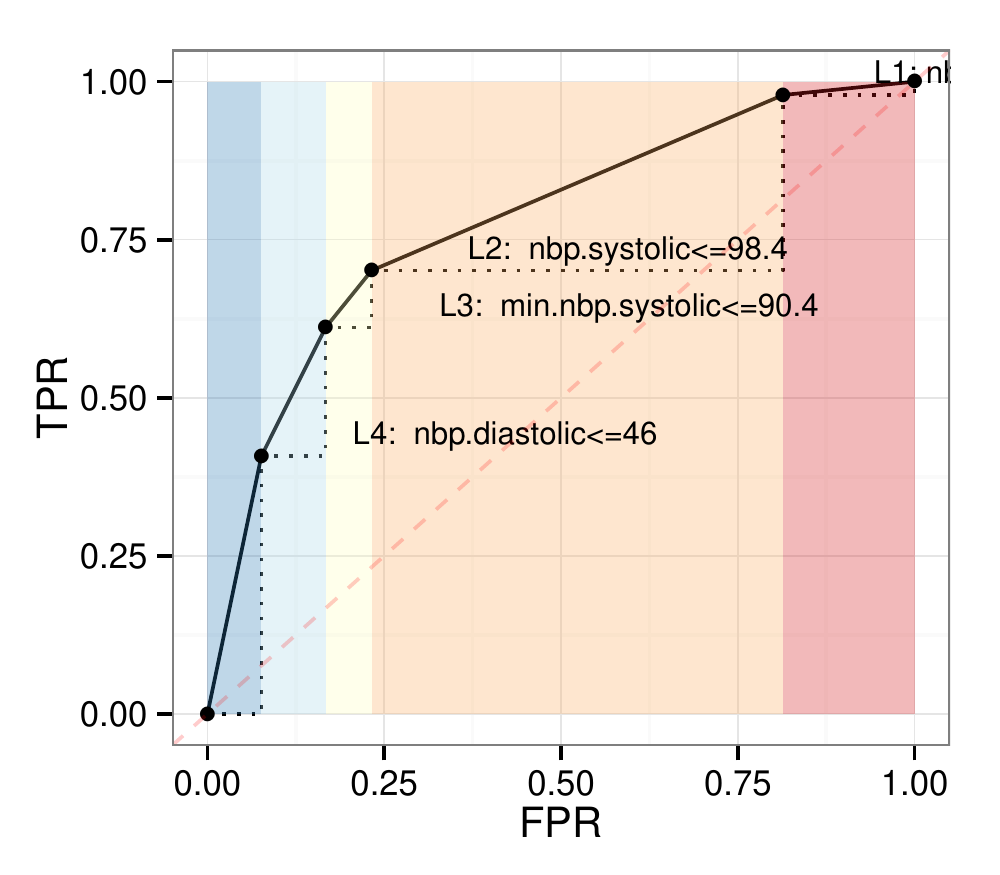}
\caption{ROC chart}
\end{subfigure}
\begin{subfigure}[b]{0.45\linewidth}
\includegraphics[width=\linewidth]{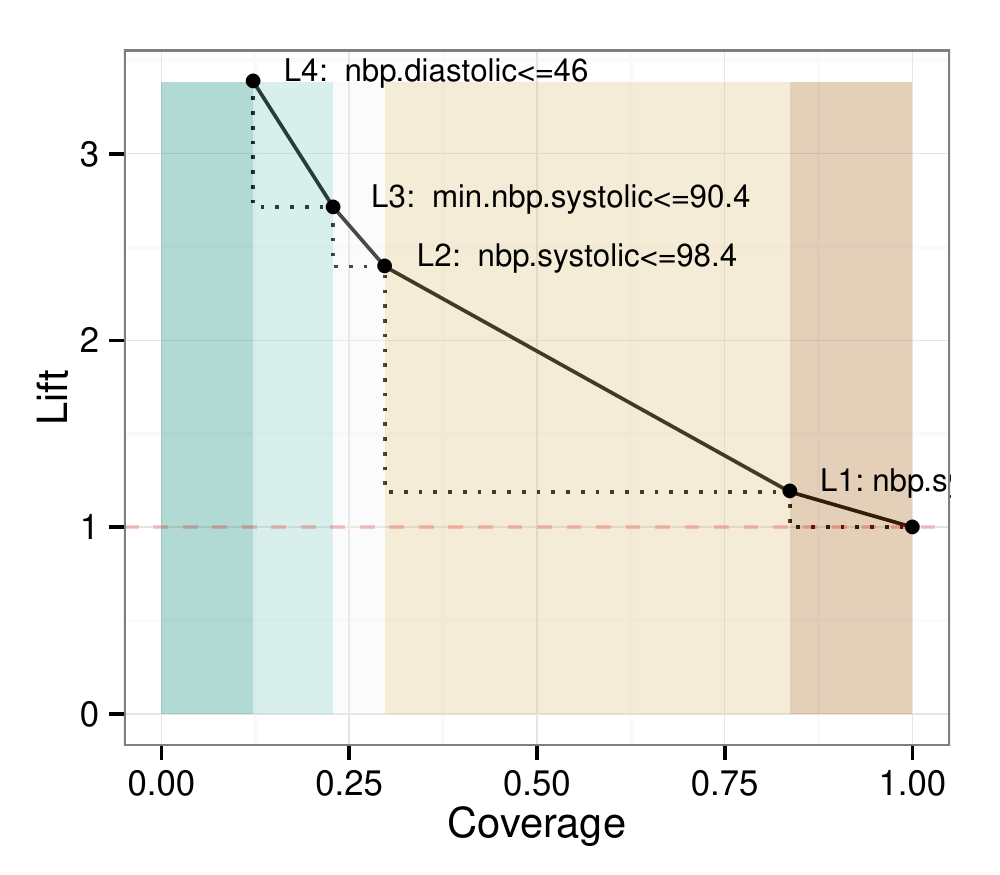}
\caption{Lift chart}
\end{subfigure}
\vspace{-10pt}
\caption{ROC curve and Lift chart. 
Every edge point on the curves is associated with a decision rule.}\label{fig:septic_roc}
\end{figure}

Figure~\ref{fig:septic_auc} (b) shows the AUC for both 1 hour and 2 hours before shock.
The decision trees are grown until they reach tree-depth 3 and ACDCs were grown until they reach chain-depth 4.
As can be seen, the performance of ACDC is statistically comparable with that of decision trees.
Figure~\ref{fig:septic_roc} shows the rule-annotated ROC curves and Lift charts.

\textbf{Cardiac Arrest.}
For the second MIMIC-II subset, we use decision trees to predict a cardiac arrest event, specifically asystole.
Cardiac arrest is a deadly condition caused by a sudden failure of heart and is often synonymous with clinical death.
Early response to cardiac arrest can reduce the mortality rate from 80\% to 60\%.
For this subset, there is a total of 3590 patients with 361 diagnosed with asystole.

\begin{figure}
\centering
\begin{subfigure}[b]{0.45\linewidth}
\includegraphics[width=\linewidth]{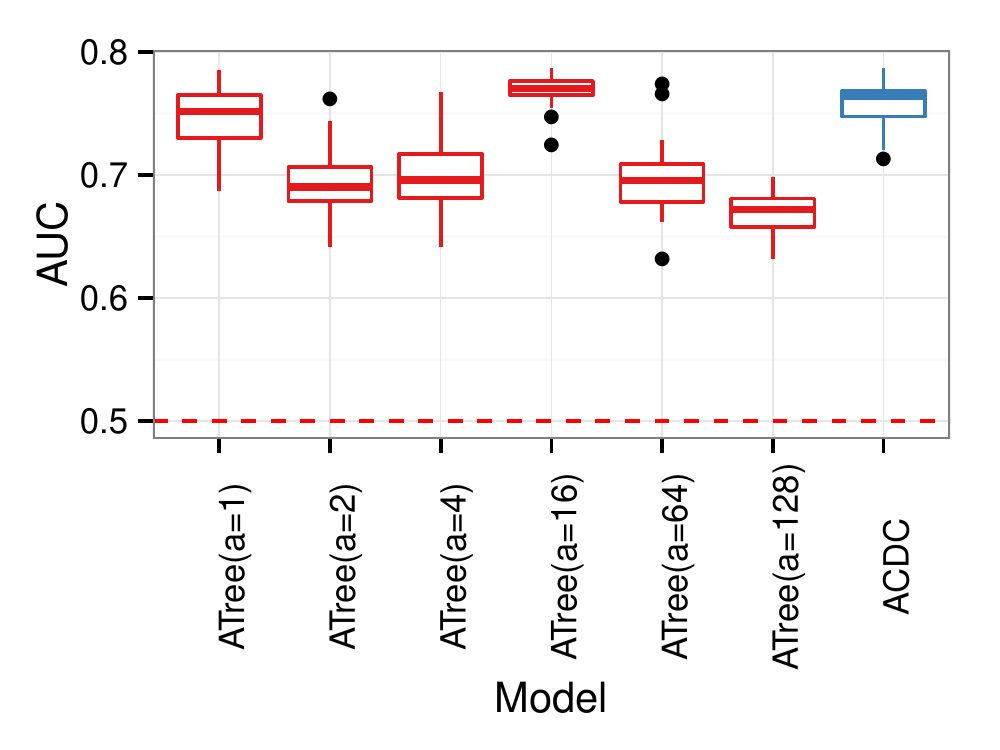}
\caption{1 hr before Asystole}
\end{subfigure}
\begin{subfigure}[b]{0.45\linewidth}
\includegraphics[width=\linewidth]{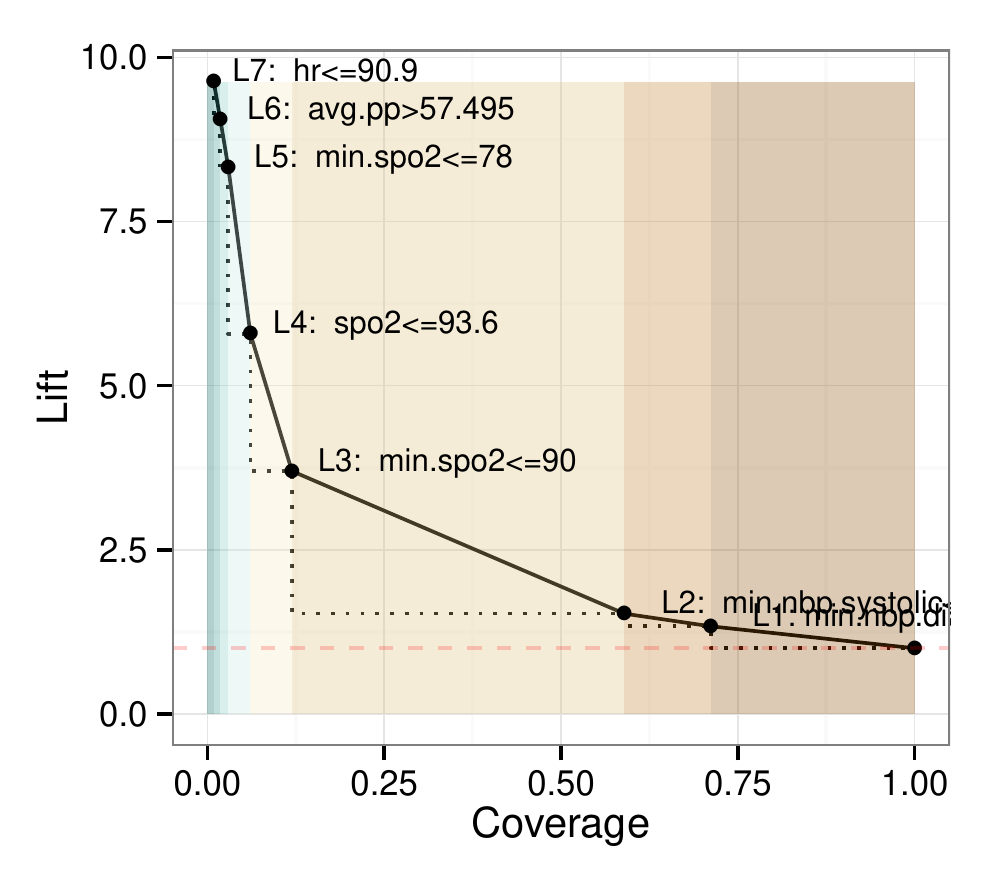}
\caption{Lift: 1hr before Asystole}
\end{subfigure}
\vspace{-10pt}
\caption{[Cardiac Arrest] The performance of ACDC is comparable to those of $\alpha$-Trees with $\alpha=1$ (C4.5) and $\alpha=2$ (CART). }
\label{fig:ca_auc}
\end{figure}

\begin{figure}[h]
\centering
\includegraphics[width=0.6\linewidth]{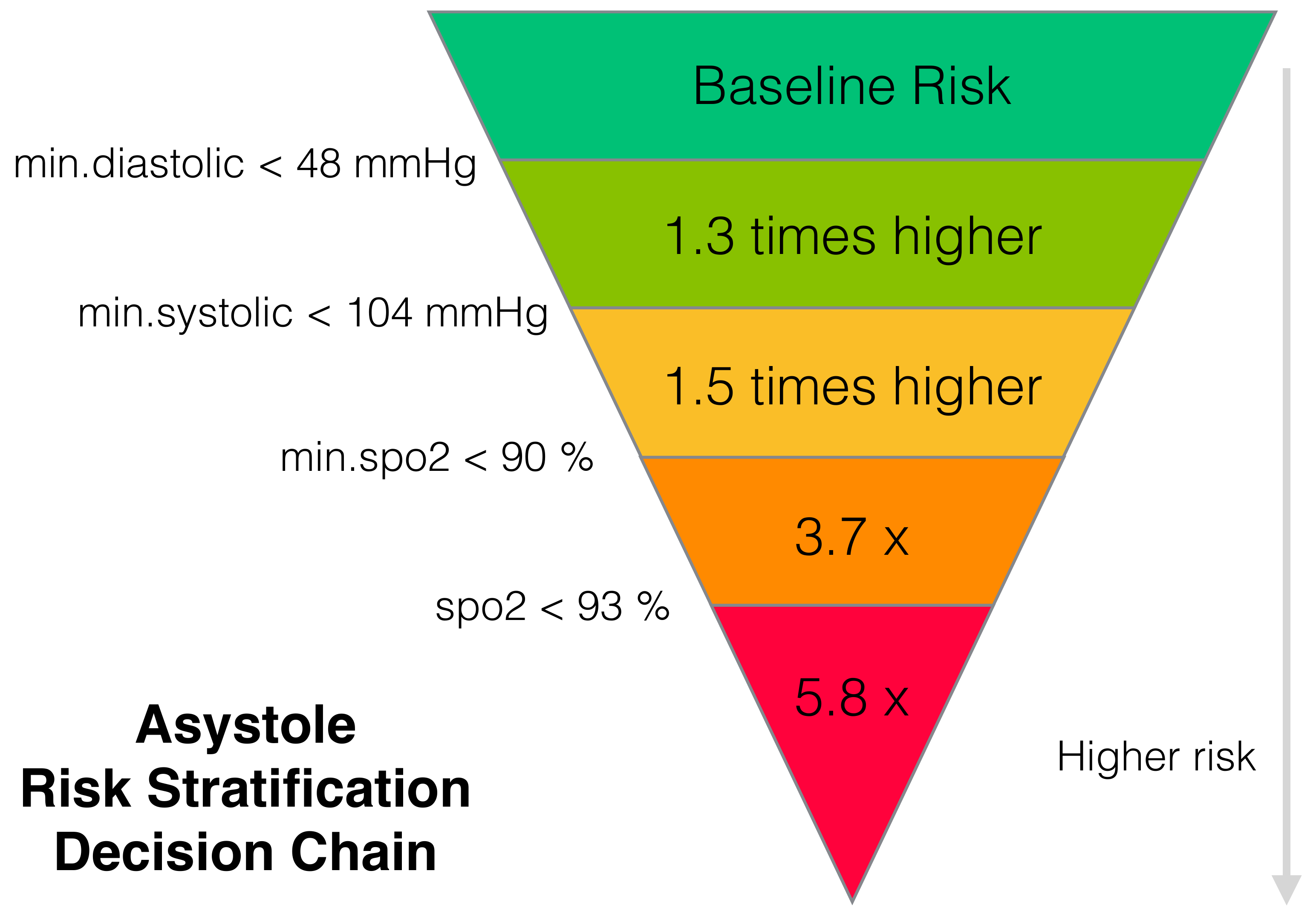}
\vspace{-10pt}
\caption{An illustrative decision chain for asystole risk stratification}
\label{fig:ca_pyramid}
\end{figure}

Figure~\ref{fig:ca_auc} (a) illustrates the predictive performance measured by AUC.
ACDC results in better predictive performance compared to the other tree-based methods.
Furthermore, the decision chains are visually interpretable.
Figure~\ref{fig:ca_auc} (b) shows the rule-annotated Lift charts.
From these charts, we can observe that asystole patients are characterized by low heart beat rates and low pulse oximetry values.
Figure~\ref{fig:ca_pyramid} illustrates the extracted decision chain visualized using a pyramid diagram.

\section{Discussions}\label{sec:discussions}

This paper introduces a novel variant of decision tree algorithms, $\alpha$-Carving Decision Chain (ACDC).
Our algorithm produces a chain of decisions to sequentially carve out ``pure" subsets of the majority class samples, leveraging two newly developed techniques: (i) ACDC splitting criterion and (ii) $\alpha$-carving strategy.
As a result, the chain of decision rules realtively leads to a pure subset of the minority class examples.

ACDC is a greedy alternative to a general framework known as Rule Lists \cite{WangRu15}.
Moreover,  our approach is particularly well-suited when exploring large and class-imbalanced datasets. 
While a decision chain may seem too restrictive, our empirical results suggest that a chain structure achieves almost similar predictive performance as normal trees in many cases.
Moreover, the resulting chain of decisions provide an intuitive interpretation in conjunction with visual performance metrics such as ROC curve and Lift chart.

\bibliography{thesis}
\bibliographystyle{icml2016}

\end{document}